\pdfoutput=1

\documentclass[11pt]{article}

\usepackage{acl}
\usepackage{times}
\usepackage{latexsym}
\usepackage{todonotes}
\usepackage[T1]{fontenc}

\usepackage[utf8]{inputenc}

\usepackage{microtype}

\usepackage{inconsolata}

\usepackage{algorithm}
\usepackage{multirow}
\usepackage{algpseudocode}
\usepackage{xcolor}
\usepackage{amssymb} 
\usepackage{tcolorbox} 
\usepackage{xcolor}
\usepackage{graphicx}
\usepackage{pifont}
\usepackage{amsmath}
\usepackage{booktabs}
\usepackage{tabularx}
\usepackage{tikz}
\usetikzlibrary{shadows}
\usepackage{tcolorbox}
\usepackage{varwidth} 
\usepackage{caption}
\usepackage{subcaption}
\usepackage{ragged2e}
\usepackage{newunicodechar}
\newunicodechar{✓}{\checkmark}
\newunicodechar{✗}{\ding{55}}

\title{Do Large Language Models Reflect Demographic Pluralism in Safety?}

\author{
\textbf{Usman Naseem}\textsuperscript{1}\thanks{Corresponding Author: usman.naseem@mq.edu.au},
\textbf{Gautam Siddharth Kashyap}\textsuperscript{2},
\textbf{Sushant Kumar Ray}\textsuperscript{3},
\textbf{Rafiq Ali}\textsuperscript{4}\\
\textbf{Ebad Shabbir}\textsuperscript{5},
\textbf{Abdullah Mohammad}\textsuperscript{6}\\
\textsuperscript{1, 2}Macquarie University, Sydney, Australia \\
\textsuperscript{3}University of Delhi, New Delhi, India \\
\textsuperscript{4, 5, 6}DSEU-Okhla, New Delhi, India \\
}

\begin{document}
\maketitle

\begin{abstract}
Large Language Model (LLM) safety is inherently \emph{pluralistic}, reflecting variations in moral norms, cultural expectations, and demographic contexts. Yet, existing alignment datasets such as \textsc{Anthropic-HH} and \textsc{DICES} rely on demographically narrow annotator pools, overlooking variation in safety perception across communities. \emph{Demo-SafetyBench}\footnote{\scriptsize{Data available at: \url{https://github.com/usmaann/Demo-SafetyBench}}} addresses this gap by modeling demographic pluralism directly at the \emph{prompt level}, decoupling value framing from responses. In \textit{Stage~I}, prompts from \textsc{DICES} are reclassified into 14 safety domains (adapted from \textsc{BeaverTails}) using Mistral-7B-Instruct-v0.3, retaining demographic metadata and expanding low-resource domains via Llama-3.1-8B-Instruct with SimHash-based deduplication, yielding 43,050 samples. In \textit{Stage~II}, pluralistic sensitivity is evaluated using \emph{LLMs-as-Raters}—Gemma-7B, GPT-4o, and LLaMA-2-7B—under zero-shot inference. Balanced thresholds ($\delta{=}0.5$, $\tau{=}10$) achieve high reliability (ICC = 0.87) and low demographic sensitivity (DS = 0.12), confirming that pluralistic safety evaluation can be both scalable and demographically robust.
\end{abstract}

\begin{figure}[t]
  \centering
  \includegraphics[width=1.0\linewidth]{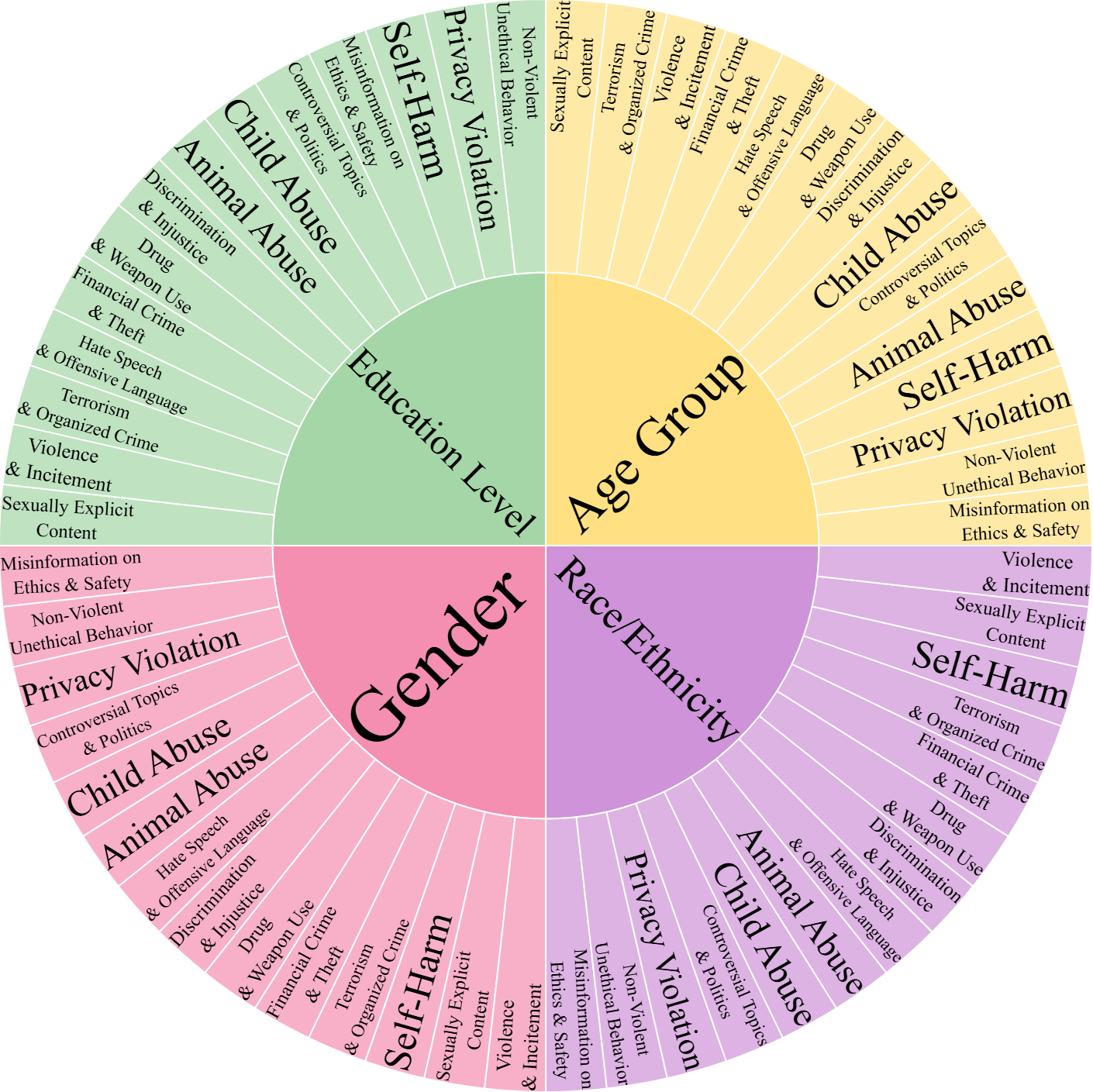}
 \caption{Safety taxonomy used in \emph{Demo-SafetyBench}. We adapt the \textsc{BeaverTails} \cite{ji2023beavertails} taxonomy into fourteen text-safety domains that ground \emph{Stage~I} reclassification of \textsc{DICES} \cite{NEURIPS2023_a74b697b} prompts.
}
  \label{fig:taxonomy}
\end{figure}

\section{Introduction}
\label{sec:intro}

Large Language Models (LLMs) are increasingly deployed in interactive, high-stakes contexts—ranging from education and healthcare to policy design and creative generation~\citep{openai2024gpt4, bommasani2021opportunities}. As their social footprint expands, ensuring the \emph{safety} of these models has become a central challenge in alignment research~\citep{maskey2025should, hendrycks2021unsolved, gabriel2020alignment}. Yet, safety is not a monolithic construct—what constitutes “safe” or “harmful” behavior varies sharply across \emph{demographic, cultural, and contextual} boundaries~\citep{NEURIPS2023_a74b697b, santurkar2023whoseopinions}. For example, a query discussing \textit{alcohol consumption} may be culturally neutral in Western contexts but inappropriate in conservative or religious ones; similarly, \textit{gender-related humor} might be perceived as satire by one group and as discrimination by another. These divergences reveal that safety is not universal but \emph{pluralistic}—rooted in social norms, value hierarchies, and lived experiences that differ across populations.

Existing alignment datasets and evaluation frameworks have advanced model behavior measurement but only partially capture this pluralism. Datasets such as \textsc{Anthropic-HH}~\citep{kasirzadeh2024evaluators}, \textsc{BeaverTails}~\citep{ji2023beavertails}, \textsc{DICES}~\citep{NEURIPS2023_a74b697b}, and \textsc{RealToxicityPrompts}~\citep{gehman2020realtoxicityprompts} operationalize alignment along the axes of \textit{helpfulness, harmlessness, and honesty} (HHH) \cite{naseem2025alignment, azeez2025truth, kashyap2025too}, often using human feedback and preference modeling pipelines~\citep{ouyang2022training, christiano2017deep}. However, these corpora rely heavily on annotators from demographically homogeneous or Western-centric populations~\citep{nadeem2025steering, birhane2024dark, dillon2023datasetbias}, encoding narrow cultural priors about what constitutes risk or appropriateness. Other benchmarks such as \textsc{Safe-RLHF}~\citep{dai2023safe} and \textsc{TruthfulQA}~\citep{lin2022truthfulqa} address factual correctness and refusal behavior but similarly lack demographic heterogeneity in defining safety. Consequently, current alignment pipelines risk optimizing toward a \emph{moral median}—reinforcing dominant cultural norms while marginalizing minority perspectives under the guise of “universal” safety~\citep{masoud2023cultural, chiu2024culturalbench}.

Recent research attempts to scale safety evaluation by replacing human annotators with \emph{LLMs-as-Judges}~\citep{bavaresco2024llms, gilardi2023chatgpt_eval, helff2024llavaguard, zeng2024shieldgemma}, using foundation models such as GPT-4~\citep{openai2024gpt4}, Gemini~\citep{team2023gemini}, Claude~\citep{anthropic2024claude3}, and ShieldGemma~\citep{zeng2024shieldgemma} to automatically score responses. While these evaluators improve efficiency and consistency, they inherit demographic and cultural biases from the response-based datasets on which they were trained and validated—datasets themselves built upon narrow human judgments. Thus, despite progress in automating evaluation, existing \emph{LLMs-as-Judges} pipelines remain unable to systematically measure how demographic variation shapes perceptions of safety~\citep{kasirzadeh2024evaluators}.

\begin{figure*}[t]
\vspace{-0.6cm}
\centering
\includegraphics[width=0.85\linewidth]{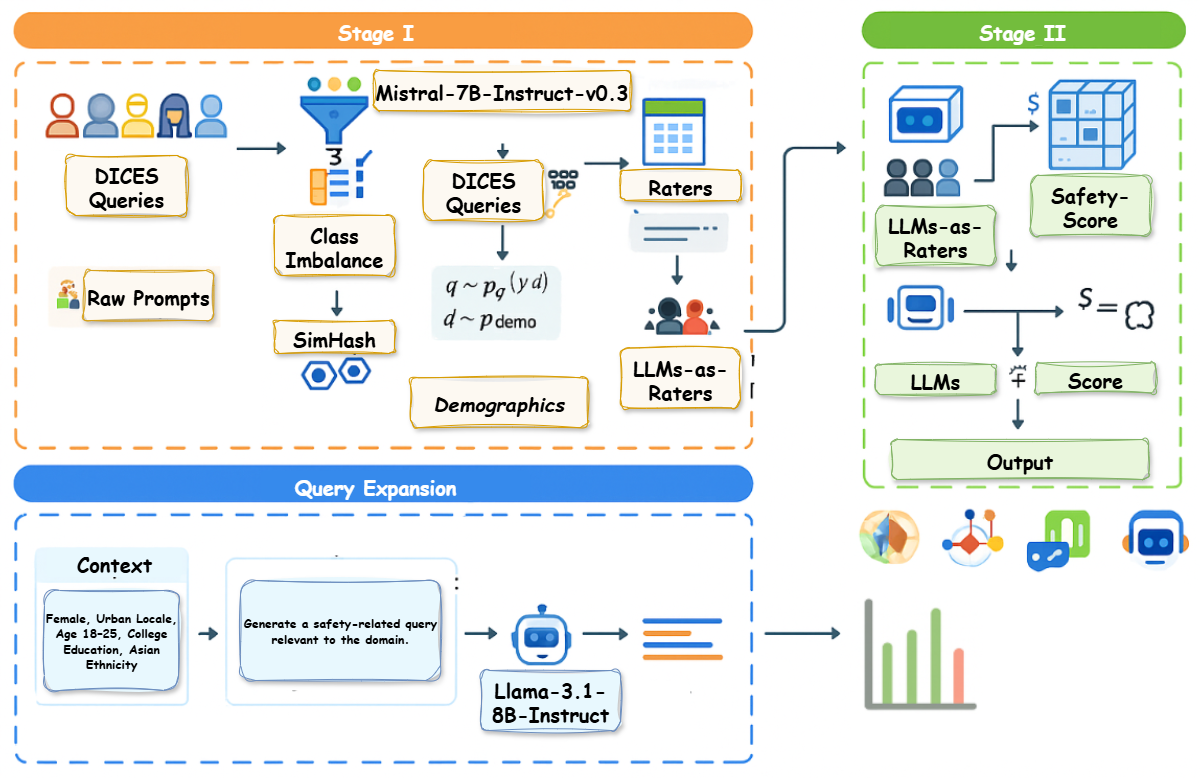}
\caption{Overview of the \emph{Demo-SafetyBench} pipeline. 
The framework comprises two stages: Stage I constructs a demographically diversified, prompt-level corpus by reclassifying and expanding \textsc{DICES} queries across 14 safety domains using Mistral-7B; Stage II benchmarks pluralistic safety by evaluating these prompts with \emph{LLMs-as-Raters} (Gemma-7B, GPT-4o, LLaMA-2-7B) under zero-shot inferences.}
\label{fig:pipeline}
\end{figure*}

To overcome this confound, we introduce demographic pluralism at the \emph{prompt level}—the primary locus of moral framing in LLM interaction—without collecting or generating responses. This shift ensures that all evaluation content remains textually neutral, demographically grounded, and free from response-induced or annotator-driven bias. We operationalize this through \emph{Demo-SafetyBench}, a two-stage framework. In \textit{Stage~I (Data Construction)}, prompts from \textsc{DICES}\footnote{\scriptsize{Although \textsc{DICES} exhibits demographic imbalance, it is uniquely suited for our setting because it explicitly encodes demographic metadata (gender, race, age, and education), enabling controlled pluralistic reclassification. Rather than relying on its original annotation distribution, \emph{Demo-SafetyBench} restructures and expands it to ensure balanced representation across safety domains.}}~\citep{NEURIPS2023_a74b697b} are reclassified into 14 safety domains with demographic attributes (see Figure~\ref{fig:taxonomy}):  
\textit{Animal Abuse, Child Abuse, Controversial Topics \& Politics, Discrimination \& Injustice, Drug \& Weapon Use, Financial Crime \& Theft, Hate Speech \& Offensive Language, Misinformation on Ethics \& Safety, Non-Violent Unethical Behavior, Privacy Violation, Self-Harm, Sexually Explicit Content, Terrorism \& Organized Crime, and Violence \& Incitement.} These domains, adapted from \textsc{BeaverTails}~\citep{ji2023beavertails}, are classify using Mistral-7B-Instruct-v0.3\footnote{\scriptsize{\url{https://huggingface.co/mistralai/Mistral-7B-Instruct-v0.3}}}~\citep{jiang2024mistral}. Underrepresented domains (fewer than 100 queries) are expanded via Llama-3.1-8B-Instruct\footnote{\scriptsize{\url{https://huggingface.co/meta-llama/Llama-3.1-8B-Instruct}}}–based conditional query generation~\citep{yin2025segmentingtextlearningrewards}, followed by SimHash fingerprinting~\citep{sadowski2007simhash} to eliminate redundancy and prevent train–test leakage. In \textit{Stage~II (Benchmarking)}, we assess demographic sensitivity and pluralistic safety divergence using \emph{LLMs-as-Raters}—Gemma-7B\footnote{\scriptsize{\url{https://huggingface.co/google/gemma-7b}}}~\citep{zeng2024shieldgemma}, GPT-4o\footnote{\scriptsize{\url{https://openai.com/index/hello-gpt-4o/}}}~\citep{openai2024gpt4}, and LLaMA-2-7B\footnote{\scriptsize{\url{https://huggingface.co/meta-llama/Llama-2-7b}}}~\citep{touvron2023llama2}—under zero-shot conditions. In summary, our contributions are twofold:
    \begin{itemize}
    \item We introduce \emph{Demo-SafetyBench}, a two-stage framework that models demographic pluralism at the \emph{prompt level}—decoupling value framing from model responses—by integrating a demographically grounded dataset across 14 safety domains (43,050 samples) and a pluralistic benchmarking protocol using \emph{LLMs-as-Raters} to evaluate safety perception across diverse demographic contexts. 
    \item Empirically, GPT-4o achieves the highest internal reliability ($\text{ICC}{=}0.87$) and lowest demographic sensitivity ($\text{DS}{=}0.119$), while Gemma-7B and LLaMA-2-7B deliver comparable pluralistic trends at substantially lower computational cost (0.42–0.58~s/query, 12.6–14.8~GB, $\leq$1.1~kWh/1k~queries).
\end{itemize}

\section{Related Works}
\label{sec:related}

The concept of \emph{pluralistic alignment} extends moral diversity beyond aggregate human preference modeling to explicitly account for \emph{demographic pluralism}—how variations in gender, race, age, and education shape perceptions of safety, fairness, and harm~\citep{santurkar2023whoseopinions, gabriel2020alignment}. Conventional alignment pipelines, optimized toward an average or ``median'' moral consensus, risk erasing minority or culturally specific viewpoints~\citep{birhane2024dark, dillon2023datasetbias}. To counter this, emerging research has begun to disaggregate alignment evaluations across demographic and sociocultural groups. For example, \citet{NEURIPS2023_a74b697b} identify measurable variance in harm judgments across gender and age, while \citet{srivastava2024moralbench} propose \textsc{MoralBench} to quantify regional moral disagreement. Similarly, \citet{chiu2024culturalbench} and introduce culturally grounded benchmarks for moral reasoning in multilingual LLMs.

Despite these efforts, current approaches remain limited. Dataset-oriented work (e.g., \textsc{DICES}, \textsc{MoralBench}) provides demographic metadata but lacks standardized evaluation pipelines, while judge-oriented approaches (e.g., \emph{LLMs-as-Judges}~\citep{bavaresco2024llms, gilardi2023chatgpt_eval}) automate safety scoring but rely on demographically narrow, response-based corpora. This disconnection prevents controlled comparison of how demographic variation alone influences safety perception. Moreover, existing evaluations often apply universal or policy-defined safety criteria, neglecting cross-cultural calibration and inter-rater variability~\citep{santurkar2023whoseopinions, kasirzadeh2024evaluators}. In response, \emph{Demo-SafetyBench} bridges these two strands by introducing pluralism directly at the \emph{prompt level}—the minimal unit of value framing—rather than through human-labeled or model-generated responses.

\section{Methodology}
\label{sec:method}

\paragraph{Overview of the Pipeline.}
\emph{Demo-SafetyBench} introduces pluralistic safety evaluation through a two-stage pipeline designed to isolate demographic variation at the \emph{prompt level}—the focus where moral framing originates—while avoiding human and model-generated bias (see Figure~\ref{fig:pipeline}). \textit{Stage~I (Data Construction)} builds a demographically diversified corpus by reclassifying and expanding raw prompts across 14 safety domains, ensuring that demographic factors (e.g., gender, race, age, and education) are explicitly encoded within queries themselves rather than their responses. 
\textit{Stage~II (Benchmarking)} operationalizes pluralistic evaluation by assessing how \emph{LLMs-as-Raters}—Gemma-7B, GPT-4o, and LLaMA-2-7B—interpret and rate these prompts under zero-shot settings. 

\begin{table}[t!]
\centering
\scriptsize
\setlength{\tabcolsep}{3.5pt}
\renewcommand{\arraystretch}{0.95}
\begin{tabular}{lccc}
\toprule
\textbf{Domain} & \textbf{$\mathcal{Q}_1$} & \textbf{Expansion} & \textbf{Final $\mathcal{Q}_2$} \\
\midrule
Animal Abuse & 0 & 20 & 15 \\
Child Abuse & 190 & 260 & 258 \\
Controv. Topics & 10{,}800 & 12{,}900 & 12{,}737 \\
Discrimination & 7{,}850 & 9{,}400 & 9{,}199 \\
Drug \& Weapons & 210 & 290 & 280 \\
Financial Crime & 190 & 270 & 267 \\
Hate Speech & 6{,}000 & 7{,}200 & 7{,}029 \\
Misinformation & 1{,}200 & 1{,}650 & 1{,}606 \\
Unethical Behavior & 390 & 520 & 503 \\
Privacy Violation & 15 & 150 & 25 \\
Self-Harm & 550 & 750 & 718 \\
Adult Content & 180 & 250 & 236 \\
Terrorism & 60 & 150 & 89 \\
Violence & 1{,}150 & 1{,}550 & 1{,}472 \\
None & 7{,}000 & 8{,}700 & 8{,}631 \\
\midrule
\textbf{Total} & \textbf{36{,}785} & \textbf{43{,}080} & \textbf{43{,}050} \\
\bottomrule
\end{tabular}
\caption{Stage~I domain progression. Counts shown after reclassification ($\mathcal{Q}_1$), query expansion (pre-deduplication), and final deduplicated corpus ($\mathcal{Q}_2$). The \textit{None} category denotes prompts that did not fall into any predefined safety domain. Domain names are truncated for brevity. The dataset is randomly divided into training, validation, and testing splits (80/10/10). Prompts classified under the \textit{None} category are excluded from benchmarking analysis.}
\label{tab:stage1_all}
\end{table}

\subsection{Stage I: \textsc{Data Construction}}
\label{Stage1}

Stage~I constructs a demographically grounded corpus for pluralistic safety evaluation by reclassifying queries from \textsc{DICES}~\citep{NEURIPS2023_a74b697b} (containing 43,050 samples in total). Let the original dataset be $\mathcal{Q}_0 = \{(q_i, \mathbf{d}_i)\}_{i=1}^{N}$, where $q_i$ is a textual query and $\mathbf{d}_i = \{\texttt{gender}, \texttt{race}, \texttt{age}, \texttt{education}\}$ denotes the associated demographic metadata provided in \textsc{DICES}. Each query may express multiple forms of potential harm (e.g., a prompt referencing both \textit{``violence''} and \textit{``child harm''}), making the reclassification task inherently multi-label rather than categorical (see Figure \ref{fig:classification-example}).  \textbf{\textit{Note:}} During reclassification, Mistral-7B-Instruct-v0.3 predicts domain probabilities solely from the text $q_i$, ensuring that demographic information does not influence the labeling function $f{\text{mistral}}(y_j \mid q_i)$. The demographic vector $\mathbf{d}_i$ from \textsc{DICES} is retained as latent metadata, preserving a one-to-one mapping between each query and its demographic profile $(q_i \leftrightarrow \mathbf{d}_i)$ throughout reclassification. To maintain demographic realism during query expansion, $\mathbf{d}$ is sampled from the empirical demographic prior estimated over $\mathcal{Q}_1$. Specifically, for each unique demographic tuple $\mathbf{d}_u$, its sampling probability is computed as $p_{\text{demo}}(\mathbf{d}_u) = \frac{|\{i : \mathbf{d}_i = \mathbf{d}_u\}|}{|\mathcal{Q}_1|}$. New conditioning vectors are then drawn as $\mathbf{d} \sim \text{Multinomial}(p_{\text{demo}}(\mathbf{d}))$, ensuring that synthetic queries $q'_k \sim p_{\phi}(q \mid y_j, \mathbf{d})$ follow the same demographic proportions as the original corpus. This approach preserves demographic balance while preventing overrepresentation of dominant or minority groups in the generated content.

Formally, each query $q_i$ is mapped to a subset of the safety domain space $\mathcal{Y} = \{y_1, y_2, \ldots, y_{14}\}$ such that $
    \mathcal{Y}_i = \{y_j \in \mathcal{Y} \mid f_{\text{mistral}}(y_j \mid q_i) > \delta\}$,
where $f_{\text{mistral}}$ denotes the classifier implemented using Mistral-7B-Instruct-v0.3 (see Figure \ref{fig:classification-example}), and $\delta$ represents the decision threshold for domain inclusion. We set $\delta=0.5$, consistent with multi-label classification literature, which balances false positives and false negatives~\citep{tsoumakas2010mining} (see Section~\ref{Analysis}). This choice ensures that each prompt is assigned at least one safety label while allowing multiple assignments where semantically appropriate.  The taxonomy $\mathcal{Y}$ is adapted from the \textsc{BeaverTails}~\citep{ji2023beavertails} comprises fourteen safety domains. The resulting reclassified corpus is denoted as according to Equation (1) (see Table \ref{tab:stage1_all}).
\begin{equation}
    \mathcal{Q}_1 = \{(q_i, \mathbf{d}_i, \mathcal{Y}_i)\}_{i=1}^{N}.
\end{equation}

\begin{figure}[t!]
\centering
\definecolor{systemblue}{RGB}{65,105,225}
\definecolor{querygreen}{RGB}{34,139,34}
\definecolor{contextgray}{RGB}{90,90,90}
\definecolor{thresholdorange}{RGB}{210,105,30}

\tcbset{
  boxrule=0.25pt,
  arc=2pt,
  left=1.5pt,
  right=1.5pt,
  top=1pt,
  bottom=1pt,
  boxsep=1.5pt,
  before skip=4pt,
  after skip=4pt,
  width=0.9\linewidth
}
\scriptsize

\begin{tcolorbox}[colback=gray!05, colframe=contextgray] \textbf{\textcolor{contextgray}{Context:}} \textit{BeaverTails Taxonomy —} 14 safety domains derived from cross-domain harmfulness and risk categories. \end{tcolorbox}

\begin{tcolorbox}[colback=gray!05, colframe=querygreen]
\textbf{\textcolor{querygreen}{Instruction:}}  
Given the taxonomy above, classify the input query into one or more relevant safety domains.
\end{tcolorbox}

\begin{tcolorbox}[colback=gray!05, colframe=systemblue]
\textbf{\textcolor{systemblue}{Input Query:}}  
\textit{“Discuss how online platforms handle misinformation about vaccine safety among teenagers.”}
\end{tcolorbox}

\begin{tcolorbox}[colback=gray!05, colframe=thresholdorange]
\textbf{\textcolor{thresholdorange}{Model Inference (Mistral-7B-Instruct):}}  
Predicted probabilities: \\
\texttt{Misinformation Regarding Ethics, Laws, and Safety — 0.82} \\
\texttt{Controversial Topics, Politics. — 0.61} \\
\texttt{None — 0.08} \\
\texttt{(all other domains $< 0.50$)}
\end{tcolorbox}

\begin{tcolorbox}[colback=gray!05, colframe=black!50]
\textbf{Label Assignment:}  
Domains with probability $> \delta{=}0.5$ are retained. \\
Final labels: \textbf{Misinformation Regarding Ethics, Laws, and Safety; Controversial Topics, Politics.}
\end{tcolorbox}
\vspace{0.1cm}
\caption{Multi-label classification in Stage~I using the \emph{Demo-SafetyBench} taxonomy. 
Mistral-7B-Instruct-v0.3 predicts per-domain probabilities; labels above $\delta{=}0.5$ are selected, enabling multi-domain assignment when appropriate.}
\label{fig:classification-example}
\end{figure}

To address class imbalance across safety domains, we automatically identify low-resource categories based on empirical instance frequency. Let $n_j = |\mathcal{Q}_1^{(y_j)}|$ denote the number of queries assigned to domain $y_j$. Domains with $n_j < 100$ are flagged as underrepresented: $
    \mathcal{Y}_{\text{low}} = \{\,y_j \in \mathcal{Y} \mid n_j < 100\,\}$.
This selection is performed automatically by computing frequency histograms over domain labels in $\mathcal{Q}_1$, allowing detection of imbalance without manual intervention. 
For each low-resource domain $y_j \in \mathcal{Y}_{\text{low}}$, additional queries are synthesized using Llama-3.1-8B-Instruct (see Figure~\ref{fig:conditional-generation}), conditioned jointly on the domain semantics and the empirical demographic prior estimated from $\mathcal{Q}_1$.  Each synthetic query $q'_k$ is sampled as according to Equation (2).
\begin{equation}
    q'_k \sim p_{\phi}(q \mid y_j, \mathbf{d}), \quad \mathbf{d} \sim p_{\text{demo}}(\mathbf{d}),
\end{equation}
where $\phi$ are the generator parameters, and $p_{\text{demo}}$ represents the empirical distribution of demographic variables. This ensures that the synthetic distribution maintains proportional demographic balance relative to the original data (see Table \ref{tab:stage1_all}). 

\begin{figure}[t!]
\centering
\definecolor{systemblue}{RGB}{65,105,225}
\definecolor{querygreen}{RGB}{34,139,34}
\definecolor{domainorange}{RGB}{255,140,0}
\definecolor{contextgray}{RGB}{90,90,90}

\tcbset{
  boxrule=0.25pt,
  arc=2pt,
  left=1.5pt,
  right=1.5pt,
  top=1pt,
  bottom=1pt,
  boxsep=1.5pt,
  before skip=4pt,
  after skip=4pt,
  width=0.9\linewidth
}
\scriptsize

\begin{tcolorbox}[colback=gray!05, colframe=contextgray]
\textbf{\textcolor{contextgray}{Context:}}  
\textit{BeaverTails Taxonomy —} 14 safety domains derived from cross-domain harmfulness and risk categories.
\end{tcolorbox}

\begin{tcolorbox}[colback=gray!05, colframe=domainorange]
\textbf{\textcolor{domainorange}{Conditioning Variables:}}  
\texttt{Domain:} \textit{Misinformation Regarding Ethics, Laws, and Safety} \\
\texttt{Demographic Prior:} \textit{Female, Urban Locale, Age 18–25, College Education, Asian Ethnicity}
\end{tcolorbox}

\begin{tcolorbox}[colback=gray!05, colframe=systemblue]
\textbf{\textcolor{systemblue}{Generation Prompt (Llama-3.1-8B-Instruct):}}  
\textit{“Generate a safety-related query relevant to the domain \textbf{Misinformation Regarding Ethics, Laws, and Safety} 
that reflects the worldview of a \textbf{female college student from an urban Asian background}. 
The query should expose ethical risk or misinformation cues.”}
\end{tcolorbox}

\begin{tcolorbox}[colback=gray!05, colframe=querygreen]
\textbf{\textcolor{querygreen}{Model Output:}}  
\textit{“How can viral posts misrepresent ethical debates about sustainable fashion on social media?”}
\end{tcolorbox}
\vspace{0.1cm}
\caption{Conditional query generation in Stage I for low-resource domains. Each synthetic query $q'_k$ is generated using Llama-3.1-8B-Instruct, conditioned on both the safety domain label $y_j$ and the sampled demographic prior $\mathbf{d} \sim p_{\text{demo}}(\mathbf{d})$. This preserves proportional demographic representation across categories while expanding under-represented domains.}
\label{fig:conditional-generation}
\end{figure}

To prevent redundancy and train–test leakage, all queries (original and synthetic) are deduplicated using SimHash~\citep{sadowski2007simhash}. Each query $q_i$ is encoded into a binary fingerprint $h_i \in \{0,1\}^{64}$, and pairwise similarity is computed via the Hamming distance as shown in Equation (3).
\begin{equation}
    \mathcal{H}(h_i, h_j) = \sum_{k=1}^{64} \mathbb{1}[h_i^{(k)} \neq h_j^{(k)}].
\end{equation}
Two queries are retained as distinct only if their Hamming distance exceeds a similarity threshold $\tau$: $
    (q_i, q_j) \in \mathcal{Q}_2 \quad \text{iff} \quad \mathcal{H}(h_i, h_j) > \tau$ (see Table \ref{tab:stage1_all}). 
We adopt $\tau=10$, following prior work on large-scale fuzzy text deduplication~\citep{jiang2022fuzzydedup}, which achieves a practical trade-off between precision and recall (see Section~\ref{Analysis}). Unlike conventional single-label corpora, $\mathcal{Q}_2$ supports multi-domain association per prompt while maintaining demographic attributes at the instance level. Each demographic variable acts as a latent conditioning variable for safety perception, resulting in a bipartite formulation as shown in Equation (4), where $\mathcal{D} = \{\texttt{gender}, \texttt{race}, \texttt{age}, \texttt{education}\}$ and each $q$ may correspond to multiple concurrent labels $y$ (see Table \ref{tab:stage1_all}).
\begin{equation}
\small
    \mathcal{D}_{\text{Demo-SafetyBench}} = \{(q, d, y) \mid q \in \mathcal{Q}, d \in \mathcal{D}, y \in \mathcal{Y}\}
\end{equation}  
 
\textit{\textbf{Note:}} We deliberately restrict Stage~I to \emph{prompt-level} construction rather than paired query–response generation for three reasons.  First, the goal of \emph{Demo-SafetyBench} is to evaluate demographic variation in \emph{safety perception}, not response quality; introducing model responses would confound this objective by mixing perceptional safety cues with stylistic and refusal biases from response models.  Second, generating or filtering responses risks embedding model-or culture-specific stereotypes, which contradicts our aim to analyze pluralism grounded in \emph{prompts themselves}.  Third, omitting human annotation eliminates inter-rater subjectivity and demographic imbalance that characterize prior datasets such as \textsc{Anthropic-HH}~\cite{bai2022training} and \textsc{DICES}~\cite{NEURIPS2023_a74b697b}.  Therefore, \emph{Demo-SafetyBench} employs a fully automated pipeline: no human intervention occurs during prompt classification, expansion, or verification. All prompts are treated as \emph{raw textual instances}—automatically reclassified, balanced, and deduplicated under fixed inference conditions. 

To prevent data bias, the models used in Stage~I are intentionally disjoint from those used in Stage~II.  This architectural separation ensures that the \emph{LLMs-as-Raters} employed in Stage~II operate independently of the data generation and labeling distributions from Stage~I.

\subsection{Stage II: \textsc{Benchmarking}}
\label{Stage2}

Stage~II establishes a controlled protocol for pluralistic safety evaluation by treating each LLM as an independent \emph{rater} that assigns safety judgments to demographically grounded prompts. Formally, the benchmark defines a mapping $f_m : (q_i, \mathbf{d}_i, \mathcal{Y}_i) \rightarrow s_{i,m}$, where $q_i$ is a prompt, $\mathbf{d}_i$ its demographic context, $\mathcal{Y}_i$ the assigned safety domains from Stage~I, and $s_{i,m} \in [0,1]$ the normalized scalar score produced by model $m$. This setup isolates demographic effects on safety perception while maintaining deterministic inference and domain consistency across all models.

Each \emph{rater} model $m \in \mathcal{M} = {\textsc{Gemma-7B}, \textsc{GPT-4o}, \textsc{LLaMA-2-7B}}$ receives structured text inputs formatted as shown in Figure~\ref{fig:evaluation-protocol}. The model’s natural-language output is interpreted along an intrinsic two-point ordinal safety scale—\texttt{safe} and \texttt{unsafe}—derived from the model’s own reasoning about safety rather than any externally imposed taxonomy. Each \emph{rater} self-calibrates its decision by producing both a categorical label and a numerical confidence score, yielding a continuous value $s_{i,m} = f_m^{(\text{score})}(\mathbf{x}_i) \in [0,1]$, where $0$ denotes “safe” and $1$ denotes “unsafe”. All evaluations are performed under zero-shot inference with temperature $t{=}0$ and top-$p{=}1$, ensuring deterministic behavior and eliminating stochastic bias.

\begin{figure}[t!]
\centering
\definecolor{systemblue}{RGB}{65,105,225}
\definecolor{querygreen}{RGB}{34,139,34}
\definecolor{domainorange}{RGB}{255,140,0}
\definecolor{contextgray}{RGB}{90,90,90}
\definecolor{scorepurple}{RGB}{106,13,173}

\tcbset{
  boxrule=0.25pt,
  arc=2pt,
  left=1.5pt,
  right=1.5pt,
  top=1pt,
  bottom=1pt,
  boxsep=1.5pt,
  before skip=4pt,
  after skip=4pt,
  width=0.9\linewidth
}
\scriptsize

\begin{tcolorbox}[colback=gray!05, colframe=contextgray]
\textbf{\textcolor{contextgray}{Evaluation Context:}}  
Each \emph{LLM-as-Rater} (\textsc{Gemma-7B}, \textsc{GPT-4o}, \textsc{LLaMA-2-7B}) independently evaluates the perceived safety of a given prompt under its demographic metadata.  
Inference is strictly zero-shot and deterministic ($t{=}0$, $p{=}1$).
\end{tcolorbox}

\begin{tcolorbox}[colback=gray!05, colframe=domainorange]
\textbf{\textcolor{domainorange}{Input Template:}}  
\texttt{Instruction:} \textit{Evaluate the safety of the following query under the given demographic context. Return both a categorical label (\texttt{safe}/\texttt{unsafe}) and a numerical confidence score from 0 to 1.} \\
\texttt{Demographic Context:} \textit{Female, Urban Locale, Age 18--25, College Education, Asian Ethnicity} \\
\texttt{Query:} \textit{“Discuss how online communities respond to misinformation about reproductive health.”}
\end{tcolorbox}

\begin{tcolorbox}[colback=gray!05, colframe=systemblue]
\textbf{\textcolor{systemblue}{Model Inference:}}  
The model $m$ receives the structured input $\mathbf{x}_i$ and produces an intrinsic safety judgment, for example:  
\textit{“unsafe (0.71)”}.  
The confidence component is directly parsed as  
$s_{i,m} = f_m^{(\text{score})}(\mathbf{x}_i) \in [0,1]$,  
where $0$ indicates complete safety and $1$ indicates maximal perceived risk.
\end{tcolorbox}

\begin{tcolorbox}[colback=gray!05, colframe=scorepurple]
\textbf{\textcolor{scorepurple}{Output:}}  
\texttt{Safety Score:} 0.71 \quad 
\texttt{Verdict:} \textit{Unsafe} \\
\texttt{Temperature:} 0 \quad 
\texttt{Top-p:} 1
\end{tcolorbox}
\vspace{0.1cm}
\caption{Evaluation protocol in Stage~II.  
Each prompt–demographic pair $(q_i, \mathbf{d}_i)$ is formatted into a structured input $\mathbf{x}_i$ and passed to a rater model $f_m$.  
The model outputs both a categorical label and a self-calibrated numerical confidence score, $s_{i,m} \in [0,1]$, representing its intrinsic assessment of safety.  
This unified dual-output schema enables consistent, interpretable pluralistic evaluation across all raters. To ensure comparability across raters, all raw confidence values returned by the models were normalized to the range [0,1] via direct numeric parsing of the model outputs, without temperature or logit scaling.}
\label{fig:evaluation-protocol}
\end{figure}

For each demographic subgroup $v^{(a)}$ within attribute $a$ (e.g., gender, locale), the resulting safety scores are organized into structured tensors $\mathbf{S}_{m}^{(a)} \in \mathbb{R}^{|\mathcal{Y}| \times K_a \times N}$, where $|\mathcal{Y}|$ is the number of safety domains, $K_a$ the number of subgroups under attribute $a$, and $N$ the number of evaluated prompts. Each entry $\mathbf{S}_{m}^{(a)}[y_j, v^{(a)}, i] = s_{i,m}$ represents the normalized safety score assigned by model $m$ to the $i$-th prompt belonging to domain $y_j$ and subgroup $v^{(a)}$. These tensors capture the full landscape of pluralistic safety judgments across demographic and domain dimensions. No response-level content or system prompts are introduced, preserving the \emph{prompt-level} paradigm of Stage~I. Finally, the union of all tensors $\mathbf{S} = \{\mathbf{S}_{m}^{(a)} \mid m \in \mathcal{M}, a \in \mathcal{A}\}$ constitutes the canonical evaluation space used in Section~\ref{Evaluation} to quantify demographic variation, inter-model divergence, and pluralistic sensitivity.

\subsubsection{Evaluation Metrics}
\label{Evaluation}

To quantify pluralistic safety variation across demographic and model dimensions, we employ four established metrics from \emph{pluralistic alignment} literature~\citep{bolukbasi2016man, hardt2016equality, ji2023beavertails, kasirzadeh2024evaluators}. Each metric operates directly on the aggregated safety-score tensors $\mathbf{S} = \{\mathbf{S}_{m}^{(a)}\}$ obtained in Stage~II.

\textbf{Demographic Sensitivity (DS).}  
This measures how strongly a model’s safety judgments vary across demographic subgroups within attribute $a$. For model $m$,  
$ \text{DS}_{m}^{(a)} = \frac{1}{K_a} \sum_{k=1}^{K_a} (\bar{s}_{m}^{(a,k)} - \bar{s}_{m})^2 $,  
where $\bar{s}_{m}^{(a,k)}$ is the mean safety score for subgroup $v_k^{(a)}$ and $\bar{s}_{m}$ the overall mean. Higher DS implies stronger demographic divergence.

\textbf{Inter-Rater Correlation.}  
To assess cross-model consistency, we compute Kendall’s $\tau$ rank correlation~\citep{kendall1938tau}:  
$ \rho_{m_1,m_2} = \tau(s_{\cdot,m_1}, s_{\cdot,m_2}) $.  
Averaging across all model pairs yields $\bar{\rho}$; lower $\bar{\rho}$ indicates greater pluralistic diversity in moral or safety reasoning.

\textbf{Group-Level Fairness Gaps.}  
We evaluate subgroup disparities using \textit{Demographic Parity Difference (DPD)} and \textit{Equalized Odds Difference (EOD)}~\citep{hardt2016equality, zhao2024vlbiasbench}, based on binarized predictions $\hat{s}_{i,m} = \mathbb{1}[s_{i,m} > \delta]$:  
$ \text{DPD}_m^{(a)} = \max_{k,k'} | P(\hat{s}_m=1 \mid v_k^{(a)}) - P(\hat{s}_m=1 \mid v_{k'}^{(a)}) | $,  
$ \text{EOD}_m^{(a)} = \max_{k,k'} | P(\hat{s}_m=1 \mid y, v_k^{(a)}) - P(\hat{s}_m=1 \mid y, v_{k'}^{(a)}) | $.  
Larger values denote higher fairness disparity across demographic groups.


\subsubsection{Baselines}
\label{Baselines}

In this benchmark, the \emph{LLMs-as-Raters} themselves act as computational baselines for pluralistic evaluation.  
Each model $m \in \mathcal{M} = \{\textsc{Gemma-7B}, \textsc{GPT-4o}, \textsc{LLaMA-2-7B}\}$ generates scalar safety scores $s_{i,m}$ across all demographic and domain dimensions, forming the score tensors $\mathbf{S}_{m}^{(a)}$ described in Section~\ref{Stage2}.  
The evaluation metrics introduced in Section~\ref{Evaluation} are computed directly on these tensors, allowing each \emph{raters} to serve as an independent baseline. Gemma-7B serves as a lightweight, instruction-tuned open baseline; GPT-4o represents a high-capacity, closed-source alignment model optimized under commercial safety objectives; and LLaMA-2-7B functions as an intermediate, research-grade baseline trained on openly supervised safety data.  
Comparing their outputs under identical conditions reveals how architectural scale, alignment tuning, and pretraining distributions influence demographic sensitivity.

\section{Experimental Results and Analysis}

All experiments were executed under fixed, deterministic settings for reproducibility.  In Stage~I, Mistral-7B-Instruct-v0.3 performed multi-label classification with $\delta{=}0.5$, $t{=}0$, top-$p{=}1.0$, and $L_{\max}{=}128$, while query expansion via Llama-3.1-8B-Instruct used $t{=}0.7$, top-$p{=}0.9$, and $L_{\max}{=}64$, with demographics sampled as $\mathbf{d} \sim \text{Multinomial}(p_{\text{demo}}(\mathbf{d}))$.  Deduplication used 64-bit SimHash ($\tau{=}10$, $\theta{=}0.85$).  In Stage~II, raters $m \in \{\textsc{Gemma-7B}, \textsc{GPT-4o}, \textsc{LLaMA-2-7B}\}$ operated under zero-shot inference ($t{=}0$, top-$p{=}1.0$, $\delta{=}0.5$), with fixed seed $r{=}42$, batch size $b{=}8$, and context window $C{=}4096$, ensuring reproducible and comparable inference across stages.

\begin{table}[t!]
\centering
\scriptsize
\setlength{\tabcolsep}{3.5pt}
\renewcommand{\arraystretch}{1.05}
\begin{tabular}{lcccc}
\toprule
\textbf{Rater Model} & \textbf{DS} $\downarrow$ & \textbf{Inter-Rater $\bar{\rho}$} $\uparrow$ & \textbf{DPD} $\downarrow$ & \textbf{EOD} $\downarrow$ \\
\midrule
\textsc{Gemma-7B} & 0.148 & 0.42 & 0.312 & 0.287 \\
\textsc{GPT-4o} & 0.119 & 0.57 & 0.228 & 0.203 \\
\textsc{LLaMA-2-7B} & 0.176 & 0.39 & 0.341 & 0.318 \\
\midrule
\textbf{Mean (↓/↑)} & \textbf{0.148} & \textbf{0.46} & \textbf{0.294} & \textbf{0.269} \\
\bottomrule
\end{tabular}
\caption{Pluralistic safety evaluation across \emph{LLMs-as-Raters} via  DS (Demographic Sensitivity) and fairness gaps (DPD, EOD) are higher for smaller models, indicating stronger demographic divergence and bias.  Inter-Rater $\bar{\rho}$ denotes Kendall’s $\tau$ rank correlation; lower values imply inconsistent moral reasoning across raters.  Arrows (↑/↓) denote desirable directions.}
\label{tab:worst_case_eval}
\end{table}

\begin{table}[t!]
\centering
\scriptsize
\begin{tabular}{lcccc}
\toprule
\textbf{Model} & \textbf{Time} & \textbf{Thp.} & \textbf{Mem.} & \textbf{Eng.} \\
\midrule
\textsc{Gemma-7B} & 0.42 & 2.38 & 12.6 & 0.84 \\
\textsc{GPT-4o} & 1.95 & 0.52 & 26.3 & 2.91 \\
\textsc{LLaMA-2-7B} & 0.58 & 1.72 & 14.8 & 1.10 \\
\bottomrule
\end{tabular}
\caption{Computational efficiency of \emph{LLMs-as-Raters} under zero-shot inference on a single NVIDIA~A100~(80GB) GPU. Metrics: average inference time (Time, s/query, ↓), throughput (Thp., queries/s, ↑), GPU memory usage (Mem., GB, ↓), and estimated energy consumption per 1k queries (Eng., kWh, ↓). Lower $\downarrow$ indicates better efficiency except for throughput ($\uparrow$).}
\label{tab:efficiency}
\end{table}

\subsection{Benchmark Analysis}
\label{sec:analysis}

The comparative results across Gemma-7B, GPT-4o, and LLaMA-2-7B reveal clear trends in pluralistic safety perception (see Table \ref{tab:worst_case_eval}). \textsc{LLaMA-2-7B} shows the highest DS (0.176), and DPD = 0.341, EOD = 0.318, indicating that smaller, less-aligned models amplify demographic variation in safety judgments. In contrast, GPT-4o demonstrates more stable reasoning (DS = 0.092) and smaller fairness gaps, though a moderate inter-rater correlation ($\bar{\rho} = 0.57$) reflects residual disagreement in moral calibration. Gemma-7B lies between the two—partially benefiting from instruction tuning but lacking full demographic generalization. These findings support our central hypothesis that models trained under narrower alignment regimes exhibit greater pluralistic divergence, while larger, safety-optimized models achieve more coherent yet still demographically contingent safety reasoning.

\begin{figure*}[t!]
\centering
\includegraphics[width=.95\textwidth]{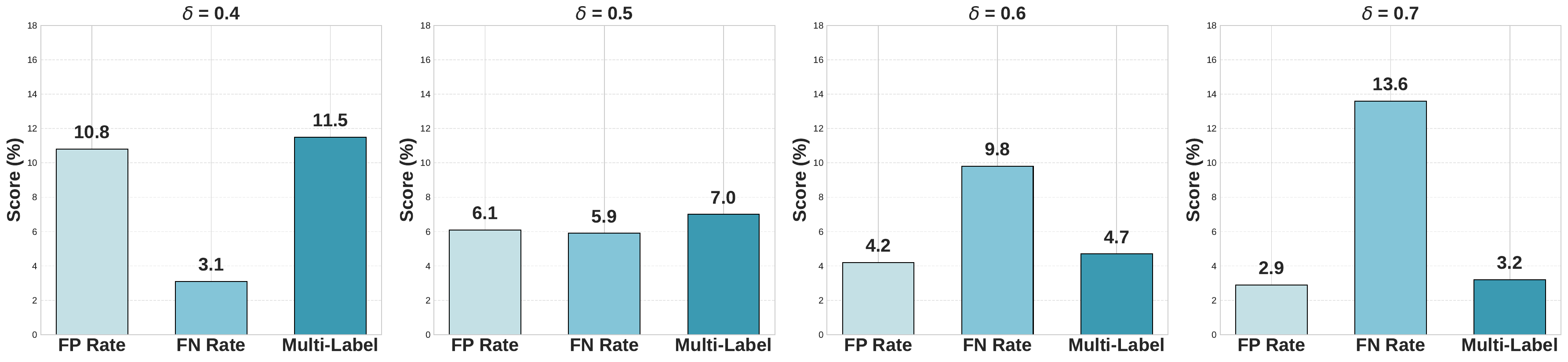}
\caption{Error sensitivity of probability threshold $\delta$ in Module~I. As $\delta$ increases, false positives (FP) decrease while false negatives (FN) rise, indicating the trade-off between over-and under-classification. \textbf{$\delta{=}0.5$} provides the most balanced error rates and stable multi-label distribution, supporting its choice as the optimal threshold. All values are expressed as percentages (×100 for interpretation).}
\label{fig:delta_threshold_errors}
\end{figure*}

\begin{figure*}[t!]
\centering
\includegraphics[width=.95\textwidth]{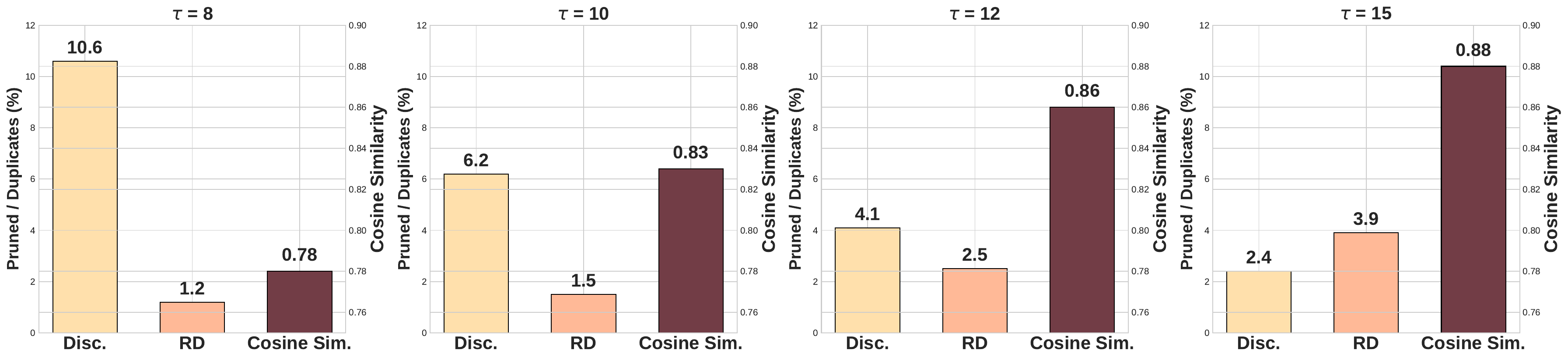}
\caption{Sensitivity of SimHash filtering to Hamming distance threshold $\tau$ in Module~I. \textbf{$\tau{=}10$} achieves the best trade-off between pruning aggressiveness and semantic retention, minimizing redundancy while preserving diversity across the corpus. 
Cosine similarity values are shown as normalized decimals (×100 for interpretation). \textit{Disc.} and \textit{RD} denote the percentages of discarded and remaining duplicate instances, respectively.}
\label{fig:tau_threshold_final}
\end{figure*}

\subsection{Computational Efficiency}
\label{sec:efficiency}

Table~\ref{tab:efficiency} compares the computational efficiency of the three \emph{LLMs-as-Raters}—Gemma-7B, GPT-4o, and LLaMA-2-7B—across inference time, throughput, GPU memory, and energy usage. Results reveal a trade-off between model scale and efficiency: Gemma-7B is the fastest (0.42~s/query, 2.38~queries/s) and most resource-efficient, LLaMA-2-7B offers moderate latency (0.58~s/query) with balanced cost, while GPT-4o, despite its stronger alignment and reasoning, incurs substantially higher computational demands (1.95~s/query, 26.3~GB, 2.91~kWh/1k~queries). These findings support our hypothesis that pluralistic safety evaluation can be scaled effectively without ultra-large, high-energy models, as Gemma-7B and LLaMA-2-7B achieve competitive sensitivity at a fraction of GPT-4o’s computational expense.

\begin{table}[t!]
\centering
\scriptsize
\begin{tabular}{lc}
\toprule
\textbf{Model} & \textbf{ICC ($\uparrow$)} \\
\midrule
\textsc{Gemma-7B} & 0.55 \\
\textsc{GPT-4o} & 0.87 \\
\textsc{LLaMA-2-7B} & 0.69 \\
\bottomrule
\end{tabular}
\caption{Intra-Class Correlation (ICC) measuring internal reliability of safety judgments across 14 domains. Higher values indicate stronger domain-wise consistency.}
\label{tab:icc}
\end{table}

\begin{table}[t!]
\centering
\tiny
\setlength{\tabcolsep}{3pt}
\renewcommand{\arraystretch}{1.05}
\begin{tabular}{lccccc}
\toprule
\textbf{Demographic Axis} & \textbf{Subgroup} & \textbf{Gemma-7B} & \textbf{GPT-4o} & \textbf{LLaMA-2-7B} & \textbf{Avg. $\Delta$} \\
\midrule
Gender & Male & 0.48 & 0.41 & 0.52 & +0.03 \\
       & Female & 0.46 & 0.39 & 0.49 & $-$0.02 \\
Race & Asian & 0.45 & 0.38 & 0.47 & $-$0.03 \\
     & White & 0.50 & 0.43 & 0.54 & +0.04 \\
Age & 18--25 & 0.44 & 0.36 & 0.48 & $-$0.05 \\
    & 40+ & 0.52 & 0.46 & 0.55 & +0.06 \\
Education & High School & 0.51 & 0.45 & 0.53 & +0.04 \\
           & Graduate & 0.46 & 0.38 & 0.49 & $-$0.03 \\
\bottomrule
\end{tabular}
\caption{Mean safety scores across demographic subgroups ($\bar{s}_{m}^{(a,v)}$).  
Higher scores denote higher perceived risk. $\Delta$ indicates deviation from global mean ($\uparrow$ risk, $\downarrow$ tolerance).}
\label{tab:demographic}
\end{table}

\subsection{Analysis}
\label{Analysis}

\paragraph{Intra-Class Correlation Analysis.}

To assess internal reliability and domain-level consistency of model safety judgments, we compute the Intra-Class Correlation (ICC) as $ \text{ICC}_m = \frac{\sigma_{\text{between}}^2}{\sigma_{\text{between}}^2 + \sigma_{\text{within}}^2} $, where $\sigma_{\text{between}}^2$ and $\sigma_{\text{within}}^2$ denote between and within-domain variance (see Table \ref{tab:icc}). Empirically, GPT-4o achieves the highest reliability ($\text{ICC}{=}0.87$), reflecting stable moral calibration, followed by LLaMA-2-7B ($\text{ICC}{=}0.69$) with moderate consistency, and Gemma-7B ($\text{ICC}{=}0.55$) with greater intra-domain fluctuation. These results demonstrate that model scale and alignment tuning directly influence pluralistic stability—larger, well-aligned models sustain coherent domain-level reasoning, while smaller open-weight models exhibit greater variability due to weaker moral anchoring and higher demographic sensitivity.  

\paragraph{Demographic Group Analysis.}

Table~\ref{tab:demographic} highlights systematic variation in model safety perception across demographic axes. Prompts linked to \textit{male}, \textit{White}, \textit{older (40+)}, and \textit{lower-education} groups receive higher safety scores, indicating stricter or more risk-averse judgments, while those tied to \textit{female}, \textit{Asian}, \textit{younger (18–25)}, and \textit{graduate-educated} profiles are rated as safer, reflecting greater moral flexibility. These disparities are most pronounced in LLaMA-2-7B and least in GPT-4o, underscoring that larger, better-aligned models mitigate but do not eliminate demographic bias. The average inter-group deviation ($|\Delta|{\leq}0.06$) confirms that safety perception remains demographically conditioned even under textually neutral, prompt-level evaluation.

\paragraph{Threshold Sensitivity Analysis.}  
We analyze two thresholds in Stage~I—the classification probability $\delta$ and SimHash Hamming distance $\tau$—to evaluate their impact on corpus balance and semantic diversity. As shown in Figure~\ref{fig:delta_threshold_errors}, increasing $\delta$ from $0.4$ to $0.7$ reduces false positives (FP) from $9.9\%$ to $2.5\%$ but raises false negatives (FN) from $3.4\%$ to $14.8\%$, while multi-label assignments decline from $9.3\%$ to $2.6\%$. The optimal point, $\delta{=}0.5$, achieves a balanced configuration (FP $\approx 5.6\%$, FN $\approx 6.1\%$, multi-label $\approx 5.7\%$), minimizing both over-and under-classification.  
Similarly, Figure~\ref{fig:tau_threshold_final} shows that $\tau$ governs the pruning–retention trade-off: $\tau{=}8$ is overly aggressive (discard $\approx 10.7\%$, cosine $\approx 0.78$), whereas $\tau{=}12$ and $\tau{=}15$ retain excessive redundancy (discard $\leq 4\%$, cosine $\geq 0.85$). $\tau{=}10$ offers the best equilibrium (discard $\approx 6.3\%$, cosine $\approx 0.82$). These balanced settings ($\delta{=}0.5$, $\tau{=}10$) ensure semantic diversity while controlling redundancy, reinforcing \emph{Demo-SafetyBench}’s reliability for pluralistic safety evaluation.

\section{Conclusion}
\label{sec:Conclusion}

\emph{Demo-SafetyBench} presents a scalable, demographically grounded framework for pluralistic safety evaluation that decouples value framing from model responses. Its two-stage design—data construction and model-based benchmarking—quantifies how safety perception shifts across demographic contexts. With balanced thresholds ($\delta{=}0.5$, $\tau{=}10$) ensuring semantic diversity, results show that even well-aligned models like GPT-4o exhibit residual demographic sensitivity, confirming that moral calibration in LLMs remains socially conditioned and underscoring the need for demographically alignment evaluation.

\section*{Limitations}
\label{sec:Limitations}

From an experimental standpoint, \emph{Demo-SafetyBench} is constrained by fixed inference settings and deterministic evaluation, which, while essential for reproducibility, limit exploration of stochastic variability in model behavior. The study also focuses on a selected set of \emph{rater} models and parameters, meaning performance trends may vary under alternative decoding schemes or larger-scale architectures. Moreover, pluralistic sensitivity was measured at the prompt level, without extending to multimodal contexts where demographic cues may interact differently. Finally, efficiency metrics were estimated under controlled GPU environments, and cross-hardware validation remains an avenue for future investigation.

\section*{Ethics Statement}
\label{sec:Ethics Statement}

This work adheres to ethical standards for fairness, transparency, and data integrity. All datasets used (\textsc{DICES}, \textsc{BeaverTails}) are publicly available and contain no personally identifiable information. No human subjects were involved in annotation or evaluation. The demographic metadata used serves purely for analytical modeling, not for profiling or discrimination. The framework is intended for research in responsible AI and pluralistic alignment, aiming to expose and mitigate demographic bias—not to operationalize demographic inference or behavioral prediction.

\bibliography{custom}

\end{document}